\documentclass[11pt]{article}
\usepackage{graphicx}
\usepackage[margin=1in]{geometry}
\usepackage{amsmath,amssymb}
\usepackage{booktabs}
\usepackage{setspace}
\usepackage{xurl}

\title{When Prediction Error Is Not Enough: Evaluating Nuisance-Function Prediction for Causal Estimation}

\author{
Cong Cao\\
Department of Biostatistics, Yale University,
New Haven, CT 06520, USA\\
Corresponding author: Cong Cao\\
Email: \texttt{cong.cao@yale.edu}
}
\date{}

\begin{document}

\maketitle
\begin{abstract}
Prediction error is widely used to evaluate nuisance-function estimators in causal inference, but its relationship with the performance of the resulting causal estimator is not well understood. We studied this question in a partially linear model using Monte Carlo simulations, comparing ordinary least squares (OLS), generalized additive models (GAMs), XGBoost, and Double Machine Learning with XGBoost (DML-XGBoost) with independent and with clustered observations. We evaluated nuisance-function prediction error, bias, RMSE, and 95\% confidence interval coverage, and derived the second-order remainder through which errors in the exposure and outcome nuisance functions jointly determine the bias of the residualized estimator. When all methods estimated the causal effect from the same training sample, DML-XGBoost had the lowest RMSE and the highest coverage among the non-oracle methods, although XGBoost fitted to the full training sample had lower out-of-sample prediction error for both nuisance functions. GAMs had the lowest prediction error of all methods but were biased, with coverage of 0.55 with independent observations. The remainder reproduced the average bias of OLS, GAMs, and DML-XGBoost, but only half to three-quarters of the bias of XGBoost, whose residuals were formed on the data used to fit it. A joint-error measure based on the absolute cross-product of the two nuisance errors was only weakly associated with causal bias. Prediction error is useful for assessing nuisance-function estimation, but it should not be treated as a direct measure of the quality of the resulting causal estimator.
\end{abstract}


\section{Introduction}

Causal effect estimation with observational data often requires estimating nuisance functions that describe the relationships between covariates, the exposure, and the outcome. Flexible regression and machine-learning methods allow these functions to be estimated without imposing restrictive parametric assumptions, and they are now widely used in causal inference \cite{hernan2020causal,kennedy2016}. Double Machine Learning (DML), for example, combines flexible nuisance-function estimation with Neyman-orthogonal scores and cross-fitting to reduce the impact of nuisance estimation error on the target causal parameter \cite{chernozhukov2018,diaz2020,zivich2021,chernozhukov2022automatic,ahrens2026}. In practice, however, nuisance models are often evaluated using the same prediction measures used in ordinary supervised learning. This raises a simple question: does better prediction of the nuisance functions also mean better causal estimation?

Prediction error is a natural way to evaluate nuisance models because estimating a nuisance function is itself a prediction problem. Cross-validation error and mean squared error are commonly used to compare learners and tune their hyperparameters \cite{hastie2009}. Recent studies have also considered whether prediction performance can help with causal analysis. Bach et al.~\cite{bach2024} studied learner choice, hyperparameter tuning, and data-splitting decisions in Double Machine Learning. They found that predictive performance can be useful for model selection, but its relationship with causal performance depends on the learner and the causal setting. Doutreligne and Varoquaux \cite{doutreligne2025} compared standard predictive criteria, including mean squared error, with criteria based more directly on causal objectives using simulations and several health-care datasets. These studies focus mainly on using prediction performance to choose models or learners. We instead ask whether the prediction error of a fitted nuisance function can be used to anticipate how well the resulting causal estimator performs. This question is important because good predictive performance does not necessarily imply good performance for a causal estimand \cite{westreich2020,naimi2023}.

The issue is more complicated when a causal estimator depends on more than one nuisance function. In the partially linear model with a continuous exposure, both the exposure regression and the outcome regression enter the DML score \cite{chernozhukov2018,chernozhukov2022automatic}. Their errors can therefore affect the causal estimator together. Orthogonality reduces the leading first-order effect of nuisance estimation error, but it does not remove finite-sample differences caused by nuisance estimation \cite{chernozhukov2018,zivich2021,ahrens2026}. We therefore examine the prediction error of each nuisance function, the second-order remainder through which the two errors jointly bias the estimator, and a simple measure of their joint error. The prediction error is measured for the nuisance functions, while bias, RMSE, and confidence interval coverage are measured for the causal estimator, so the two sets of measures capture different parts of the analysis.

We study these questions using Monte Carlo simulation in a partially linear causal model with continuous exposure. We compare ordinary least squares (OLS), generalized additive models (GAMs), XGBoost, and Double Machine Learning with XGBoost (DML-XGBoost). We evaluate nuisance-function prediction error, causal bias, RMSE, and 95\% confidence interval coverage, and we compare the bias predicted by the remainder with the observed bias. Finally, we consider a clustered-data setting with cluster-level cross-fitting to determine whether the main findings change when observations within clusters are dependent.

\section{Methods}

\subsection{Data-Generating Framework}

Let $X_i$ denote a continuous exposure, $Y_i$ the outcome, and $C_i=(C_{i1},\ldots,C_{ip})^\top$ a vector of observed baseline covariates. We generated data from the partially linear model
\[
X_i=m_0(C_i)+\epsilon_{Xi},
\qquad
Y_i=\tau X_i+g_0(C_i)+\epsilon_{Yi},
\]
where $m_0(C_i)=E(X_i\mid C_i)$ is the exposure nuisance function and $g_0(C_i)$ is the outcome baseline regression function. The causal effect $\tau$ was constant across individuals. We assumed $E(\epsilon_{Xi}\mid C_i)=0$ and $E(\epsilon_{Yi}\mid X_i,C_i)=0$. The outcome regression entering the partially linear orthogonal score is
\[
h_0(C_i)=E(Y_i\mid C_i)=\tau m_0(C_i)+g_0(C_i),
\]
so the two nuisance functions used in the causal estimator were $m_0(C)$ and $h_0(C)$. The true functions were known by construction, allowing prediction error to be evaluated directly rather than through noisy observed outcomes.

\paragraph{Simulation parameters.}
Covariates were drawn as $C_i\sim N_{10}(0,\Sigma)$ with $\Sigma_{jj}=1$ and $\Sigma_{jk}=0.3$ ($j\neq k$), with $m_0(C)=0.8C_1+0.5C_2+0.4\sin(C_3)+0.3C_4C_5$, $g_0(C)=0.7C_1-0.5C_2+0.4C_3^2+0.3C_4C_5$, $\tau=1$, and independent $\epsilon_{Xi},\epsilon_{Yi}\sim N(0,1)$. The interaction $C_4C_5$ enters both $m_0$ and $g_0$. Each replicate generated a training sample ($n=1{,}000$) and an independent evaluation sample ($n=10{,}000$); each scenario used 500 replicates. GAMs used a penalized smooth $s(C_j)$ for each covariate with REML (\texttt{mgcv}); XGBoost used learning rate 0.05, maximum depth 4, row and column subsampling 0.8, and 300 rounds, without tuning. DML used five folds.

\subsection{Nuisance-Function Error and Causal Estimation}

For fitted nuisance functions $\widehat m(C)$ and $\widehat h(C)$ we defined the prediction errors
\[
\delta_m(C)=\widehat m(C)-m_0(C),
\qquad
\delta_h(C)=\widehat h(C)-h_0(C),
\]
and summarized them by the mean squared prediction errors
\[
\mathcal{E}_m=E\{\delta_m(C)^2\},
\qquad
\mathcal{E}_h=E\{\delta_h(C)^2\}.
\]
These were computed on the evaluation sample, which was not used for fitting any nuisance model or for estimating the causal effect, so they represent out-of-sample discrepancies from the known data-generating functions. They measure the average magnitude of prediction error over the covariate distribution but contain no information about where that error occurs.

We additionally considered a simple descriptive measure of joint nuisance-function error,
\[
\mathcal{D}_{\mathrm{joint}}=\left|E\{\delta_m(C)\delta_h(C)\}\right|.
\]
The signed quantity $E\{\delta_m(C)\delta_h(C)\}$ describes the direction of the average alignment between the two nuisance-function errors, whereas $\mathcal{D}_{\mathrm{joint}}$ discards this direction. A value near zero may therefore arise either because the two errors are small or because positive and negative contributions cancel across the covariate distribution.

We compared three nuisance-function learners---OLS, GAMs, and XGBoost---using the same residualized treatment-effect estimator. OLS provided a parametric benchmark, GAMs allowed nonlinear additive relationships, and XGBoost allowed nonlinearities and interactions. For each learner, separate models were fitted for $m(C)$ and $h(C)$, and the resulting residualized exposure and outcome were used to estimate $\tau$. We additionally considered DML-XGBoost, in which the same XGBoost learner was used within a cross-fitted procedure. Because all methods use the same Neyman-orthogonal residualized estimator, the comparison between XGBoost and DML-XGBoost isolates the effect of cross-fitting.

For OLS, GAM, and XGBoost, nuisance functions were estimated on the full training sample, and the residualized variables used to estimate $\tau$ were formed on the same training observations. For DML-XGBoost, the nuisance predictions $\widehat m^{(-k)}(C_i)$ and $\widehat h^{(-k)}(C_i)$ for observation $i$ in fold $k$ were obtained from the observations outside fold $k$, and the causal effect was estimated from the cross-fitted residuals
\[
\widetilde X_i=X_i-\widehat m^{(-k(i))}(C_i),
\qquad
\widetilde Y_i=Y_i-\widehat h^{(-k(i))}(C_i),
\qquad
\widehat{\tau}=\frac{\sum_{i=1}^n\widetilde X_i\widetilde Y_i}{\sum_{i=1}^n\widetilde X_i^2}.
\]
The prediction error of DML-XGBoost was calculated by applying each fold-specific fit to the evaluation sample and averaging the squared errors across folds. The oracle estimator applied the same estimator to the training sample with the true nuisance functions.

\paragraph{Second-order remainder.} Under the model, $X_i-\widehat m(C_i)=\epsilon_{Xi}-\delta_m(C_i)$ and $Y_i-\widehat h(C_i)-\tau\{X_i-\widehat m(C_i)\}=\epsilon_{Yi}-\{\delta_h(C_i)-\tau\delta_m(C_i)\}$, so
\[
\widehat\tau-\tau=\frac{n^{-1}\sum_i\{\epsilon_{Xi}-\delta_m(C_i)\}\big[\epsilon_{Yi}-\{\delta_h(C_i)-\tau\delta_m(C_i)\}\big]}{n^{-1}\sum_i\{\epsilon_{Xi}-\delta_m(C_i)\}^2}.
\]
If the nuisance estimates do not depend on the observations at which the residuals are formed, as under cross-fitting, the terms that pair a nuisance error with $\epsilon_{Xi}$ or $\epsilon_{Yi}$ have mean zero, and
\begin{equation}\label{eq:remainder}
E(\widehat\tau)-\tau\approx\frac{R}{\mathrm{Var}(X\mid C)+\mathcal{E}_m},
\qquad
R=E\big[\delta_m(C)\{\delta_h(C)-\tau\delta_m(C)\}\big]=E\{\delta_m(C)\delta_h(C)\}-\tau\,\mathcal{E}_m.
\end{equation}
Two features of~\eqref{eq:remainder} matter here. First, the bias depends on the signed cross-product and on $\mathcal{E}_m$, whereas $\mathcal{D}_{\mathrm{joint}}$ keeps only the absolute value of the cross-product. Second, when residuals are formed on the observations used to fit the nuisance models, as for OLS, GAM, and XGBoost in this study, the nuisance errors at those observations depend on their own $\epsilon_{Xi}$ and $\epsilon_{Yi}$, and an additional own-observation term appears that errors measured on the evaluation sample do not capture. In each replicate we computed $R$ on the evaluation sample and the predicted bias $R/\{\mathrm{Var}(X\mid C)+\mathcal{E}_m\}$, with $\mathrm{Var}(X\mid C)=1$ in Scenario~1 and $1+0.8^2=1.64$ in Scenario~2.

\subsection{Clustered Setting}

To evaluate the sensitivity of the results to within-cluster dependence, we also generated
\[
X_i=m_0(C_i)+u^X_{j(i)}+\epsilon_{Xi},
\qquad
Y_i=\tau X_i+g_0(C_i)+u^Y_{j(i)}+\epsilon_{Yi},
\]
where $j(i)$ denotes the cluster containing individual $i$ and $u^X_j$ and $u^Y_j$ are cluster-level random effects, independent of the covariates and individual-level errors, with mean zero and standard deviation 0.8. Training and evaluation samples were organized into clusters of size 10, and the individual-level error standard deviations were 1.0. Five-fold cross-fitting was performed at the cluster level, so that all observations from a cluster were assigned to the same fold. This scenario was treated as a secondary analysis rather than a systematic evaluation of clustering strength.

\subsection{Simulation Design and Performance Evaluation}

Scenario~1 used independent observations and Scenario~2 the clustered setting. The nonlinear components of $m_0(C)$ and $g_0(C)$ were fixed across replicates, so that differences across methods reflected estimation performance rather than changes in the data-generating mechanism, and the same simulated data were used for all methods within a replicate. For each method and scenario we recorded $\mathcal{E}_m$, $\mathcal{E}_h$, $\mathcal{D}_{\mathrm{joint}}$, the remainder-predicted bias, bias, RMSE, empirical and estimated standard errors, and confidence interval coverage, with
\[
\mathrm{Bias}=E(\widehat{\tau})-\tau,
\qquad
\mathrm{RMSE}=\left[E\{(\widehat{\tau}-\tau)^2\}\right]^{1/2}.
\]
Coverage was the proportion of replicates in which the nominal 95\% confidence interval contained $\tau$.

\subsection{Inference}

Inference was based on the estimated influence function of the orthogonal moment. With $\widetilde X_i$ and $\widetilde Y_i$ the residuals used to estimate $\tau$, let
\[
\widehat\psi_i=\widetilde X_i\left\{\widetilde Y_i-\widehat\tau\widetilde X_i\right\}.
\]
The empirical variance of $\widehat\psi_i$ was divided by the squared derivative of the estimating equation, $(n^{-1}\sum_i\widetilde X_i^2)^2$, to obtain the standard error of $\widehat\tau$, and Wald-type 95\% confidence intervals were constructed as $\widehat\tau\pm1.96\,\widehat{\mathrm{SE}}(\widehat\tau)$. In the clustered setting, the variance was estimated from cluster-level sums of $\widehat\psi_i$ with the finite-sample factor $G/(G-1)$, where $G$ is the number of clusters.

\section{Results}

\begin{table}[htbp]
\centering
\caption{Causal estimation performance across simulation scenarios.}
\label{tab:causal_performance}
\small
\begin{tabular}{llrrrrrr}
\toprule
Scenario & Method & Bias & RMSE & Empirical SE & Mean model SE & SE ratio & Coverage \\
\midrule
1 & Oracle      &  0.0004 & 0.0310 & 0.0310 & 0.0316 & 1.018 & 0.952 \\
1 & OLS         &  0.1073 & 0.1139 & 0.0382 & 0.0380 & 0.995 & 0.192 \\
1 & GAM         &  0.0575 & 0.0658 & 0.0320 & 0.0318 & 0.996 & 0.550 \\
1 & XGBoost     & -0.0115 & 0.0430 & 0.0415 & 0.0357 & 0.859 & 0.884 \\
1 & DML-XGBoost & -0.0057 & 0.0372 & 0.0368 & 0.0333 & 0.904 & 0.928 \\
\midrule
2 & Oracle      &  0.0009 & 0.0482 & 0.0482 & 0.0478 & 0.991 & 0.948 \\
2 & OLS         &  0.0693 & 0.0853 & 0.0498 & 0.0486 & 0.976 & 0.702 \\
2 & GAM         &  0.0349 & 0.0590 & 0.0477 & 0.0461 & 0.966 & 0.858 \\
2 & XGBoost     & -0.0141 & 0.0568 & 0.0551 & 0.0464 & 0.844 & 0.892 \\
2 & DML-XGBoost & -0.0105 & 0.0522 & 0.0511 & 0.0450 & 0.879 & 0.908 \\
\bottomrule
\end{tabular}

\vspace{0.15cm}

\begin{minipage}{0.95\textwidth}
\small
\textit{Note.}
All estimators used the training sample ($n=1{,}000$); 500 replicates per scenario. Bias is the Monte Carlo mean of $\widehat{\tau}-\tau$; RMSE is the root mean squared error. Empirical SE is the standard deviation of $\widehat{\tau}$ across replicates; mean model SE is the average estimated standard error; SE ratio is mean model SE divided by empirical SE. Coverage is the proportion of nominal 95\% confidence intervals containing $\tau$ (Monte Carlo SE at most 0.022). Scenario~2 uses cluster-level cross-fitting and cluster-robust variance estimation.
\end{minipage}
\end{table}

Table~\ref{tab:causal_performance} summarizes causal estimation performance. In both scenarios, DML-XGBoost had the lowest RMSE among the non-oracle methods (0.0372 in Scenario~1 and 0.0522 in Scenario~2), followed by XGBoost (0.0430 and 0.0568), GAM (0.0658 and 0.0590), and OLS (0.1139 and 0.0853). DML-XGBoost also had the highest coverage among these methods (0.928 and 0.908, compared with 0.884 and 0.892 for XGBoost). OLS and GAM were substantially biased (0.107 and 0.058 in Scenario~1), with coverage of 0.192 and 0.550. Their standard errors were accurate (SE ratio 0.97 to 1.00), so their undercoverage came from bias. The biases of XGBoost and DML-XGBoost were small ($-0.011$ and $-0.006$ in Scenario~1), and their undercoverage came mainly from standard errors that were too small (SE ratio 0.84 to 0.90). The oracle estimator had bias 0.0004 and coverage 0.952 in Scenario~1 and bias 0.0009 and coverage 0.948 in Scenario~2.

\begin{table}[htbp]
\centering
\caption{Nuisance-function prediction error, remainder-predicted bias, and observed bias.}
\label{tab:prediction}
\small
\begin{tabular}{llrrrrrrr}
\toprule
Scenario & Method & $\mathcal{E}_m$ & $\mathcal{E}_h$ & $\mathcal{D}_{\mathrm{joint}}$ & Predicted bias & Observed bias & $r_{\mathrm{pred}}$ & $r_{\mathcal{D}}$ \\
\midrule
1 & OLS         & 0.121 & 0.846 & 0.243 & 0.1088 & 0.1073 & 0.02 & 0.03 \\
1 & GAM         & 0.094 & 0.331 & 0.163 & 0.0628 & 0.0575 & 0.02 & -0.03 \\
1 & XGBoost     & 0.169 & 0.412 & 0.163 & -0.0055 & -0.0115 & 0.03 & -0.05 \\
1 & DML-XGBoost & 0.191 & 0.470 & 0.186 & -0.0046 & -0.0057 & 0.24 & -0.05 \\
\midrule
2 & OLS         & 0.133 & 0.866 & 0.254 & 0.0684 & 0.0693 & -0.06 & -0.04 \\
2 & GAM         & 0.112 & 0.370 & 0.180 & 0.0387 & 0.0349 & -0.01 & 0.04 \\
2 & XGBoost     & 0.250 & 0.583 & 0.231 & -0.0101 & -0.0141 & 0.25 & -0.08 \\
2 & DML-XGBoost & 0.283 & 0.664 & 0.264 & -0.0100 & -0.0105 & 0.39 & -0.09 \\
\bottomrule
\end{tabular}

\vspace{0.15cm}

\begin{minipage}{0.95\textwidth}
\small
\textit{Note.}
Means over 500 replicates. $\mathcal{E}_m$, $\mathcal{E}_h$, and $\mathcal{D}_{\mathrm{joint}}$ are computed on the evaluation sample. Predicted bias is the mean of $R/\{\mathrm{Var}(X\mid C)+\mathcal{E}_m\}$ from~\eqref{eq:remainder}; observed bias is the mean of $\widehat\tau-\tau$ (Monte Carlo SE 0.0014 to 0.0025). $r_{\mathrm{pred}}$ is the correlation across replicates between predicted bias and $\widehat\tau-\tau$; $r_{\mathcal{D}}$ is the correlation between $\mathcal{D}_{\mathrm{joint}}$ and $|\widehat\tau-\tau|$.
\end{minipage}
\end{table}

\begin{figure}[htbp]
    \centering
    \includegraphics[width=\textwidth]{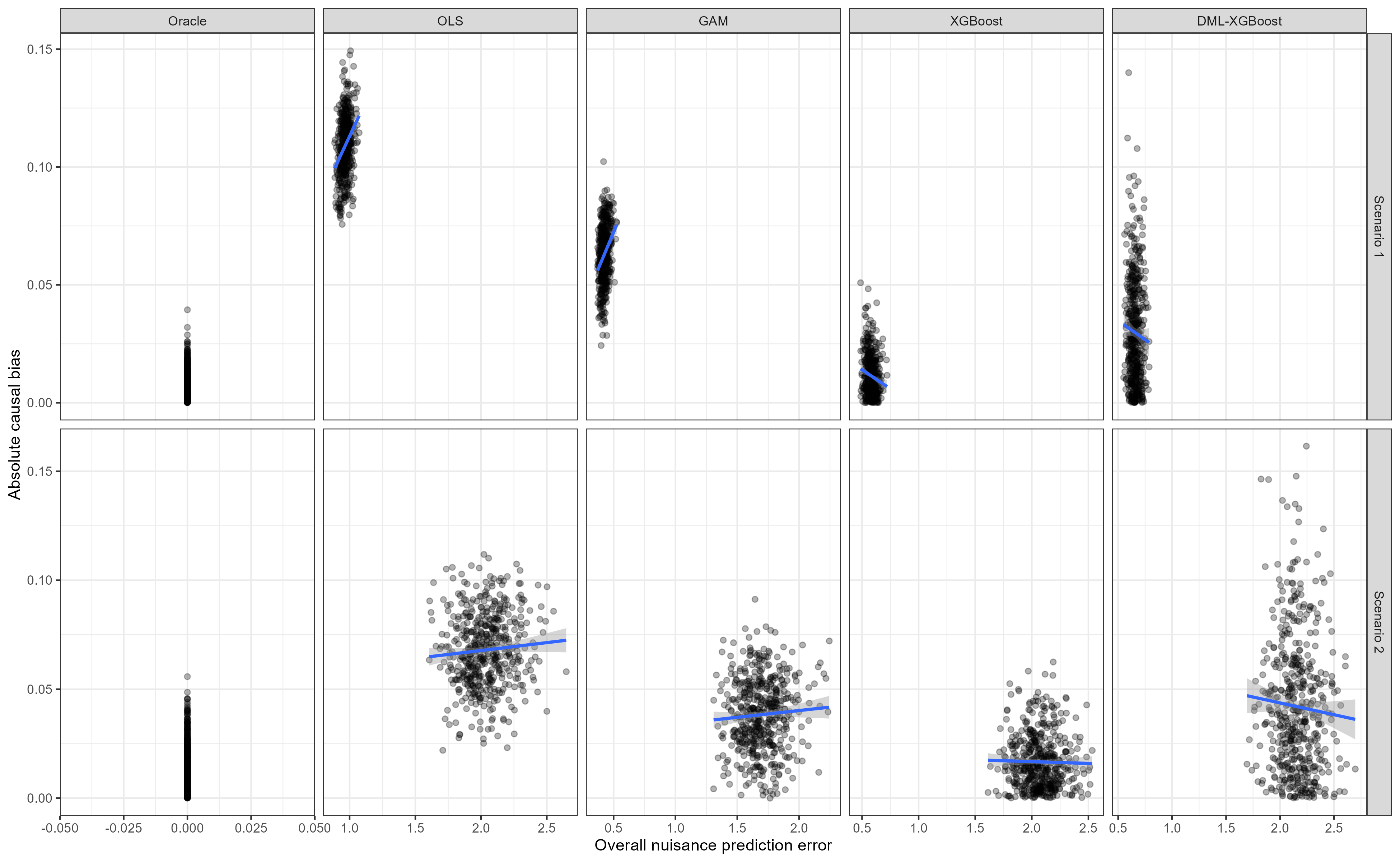}
    \caption{Total nuisance-function prediction error ($\mathcal{E}_m+\mathcal{E}_h$) and absolute estimation error $|\widehat\tau-\tau|$ across Monte Carlo replicates, by method and scenario. Lines are least-squares fits with 95\% confidence bands.}
    \label{fig:prediction-error-causal-bias}
\end{figure}

Ranking the methods by nuisance prediction error did not reproduce their ranking by causal performance (Table~\ref{tab:prediction}). GAM had the lowest prediction error for both nuisance functions in both scenarios ($\mathcal{E}_m=0.094$ and $\mathcal{E}_h=0.331$ in Scenario~1), yet it had the second-largest bias. XGBoost had lower prediction error than DML-XGBoost for both functions (0.169 and 0.412 compared with 0.191 and 0.470 in Scenario~1; 0.250 and 0.583 compared with 0.283 and 0.664 in Scenario~2) but a higher RMSE and lower coverage. OLS had a lower $\mathcal{E}_m$ than either XGBoost estimator and the largest bias. Within methods, total prediction error $\mathcal{E}_m+\mathcal{E}_h$ was at most weakly associated with absolute estimation error (Figure~\ref{fig:prediction-error-causal-bias}).

The remainder explains these discrepancies. OLS and GAM cannot represent the interaction $C_4C_5$, which enters both $m_0$ and $h_0$, so their errors in the two functions are aligned and the cross-product $E(\delta_m\delta_h)$ is large even when each error is small. The predicted bias matched the observed bias to within about 10\% for OLS (0.109 predicted and 0.107 observed in Scenario~1; 0.068 and 0.069 in Scenario~2) and GAM (0.063 and 0.058; 0.039 and 0.035), and to within Monte Carlo error for DML-XGBoost ($-0.0046$ and $-0.0057$; $-0.0100$ and $-0.0105$). For XGBoost, the predicted bias was $-0.0055$ and $-0.0101$, about half and three-quarters of the observed $-0.0115$ and $-0.0141$. This is the own-observation term: XGBoost residuals were formed on the data used to fit it, and its flexible fit absorbs part of the noise in those observations, which errors measured on the evaluation sample do not reflect. Cross-fitting removes this term, which is why the same learner had a smaller bias and a remainder that matched it. Across replicates, the predicted bias explained little of the variation in $\widehat\tau-\tau$ (correlations of 0.24 and 0.39 for DML-XGBoost and at most 0.25 for the other methods), because that variation is dominated by the first-order sampling term, which has mean zero.

\begin{figure}[htbp]
    \centering
    \includegraphics[width=\textwidth]{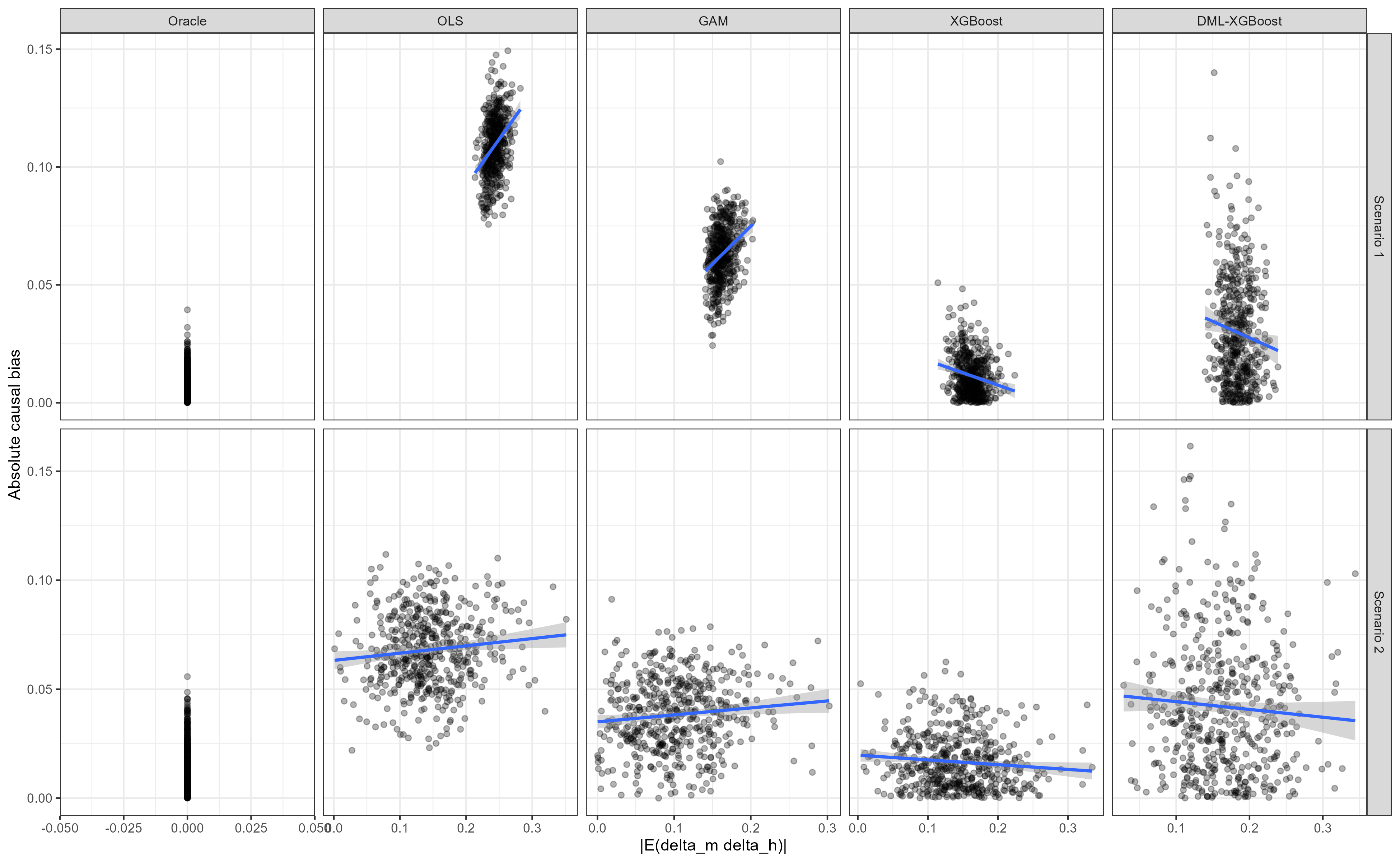}
    \caption{Joint nuisance-error measure $\mathcal{D}_{\mathrm{joint}}=|E(\delta_m\delta_h)|$ and absolute estimation error $|\widehat\tau-\tau|$ across Monte Carlo replicates, by method and scenario. Lines are least-squares fits with 95\% confidence bands.}
    \label{fig:joint-error-causal-bias}
\end{figure}

The joint-error measure $\mathcal{D}_{\mathrm{joint}}$ was weakly associated with absolute estimation error: within methods, the correlations ranged from $-0.05$ to $0.03$ in Scenario~1 and from $-0.09$ to $0.04$ in Scenario~2 (Figure~\ref{fig:joint-error-causal-bias}). This is consistent with~\eqref{eq:remainder}, since $\mathcal{D}_{\mathrm{joint}}$ discards the sign of the cross-product and omits the term $-\tau\mathcal{E}_m$. We therefore view $\mathcal{D}_{\mathrm{joint}}$ as a descriptive diagnostic of the alignment of the two nuisance errors, not as a criterion for ranking or selecting causal estimators.

The clustered setting gave the same pattern. The biases of OLS and GAM were smaller than in Scenario~1 (0.069 and 0.035), as~\eqref{eq:remainder} predicts from the larger $\mathrm{Var}(X\mid C)$ in its denominator, and all estimators were more variable. The cluster-robust standard errors were accurate for the oracle estimator (SE ratio 0.99). Because only one cluster size and one set of variance components were considered, this analysis is a sensitivity analysis rather than a systematic study of clustering strength.

\section{Discussion}

In this simulation, nuisance prediction error did not rank the methods by causal performance. The method with the lowest prediction error for both nuisance functions, GAM, was among the most biased, and XGBoost fitted to the full training sample predicted both nuisance functions better than its cross-fitted counterpart but estimated the causal effect less accurately and with lower coverage. Because all methods used the same orthogonal score, the comparison between XGBoost and DML-XGBoost isolates cross-fitting: forming residuals on held-out folds removed an own-observation term that the prediction error of a fitted function cannot reveal.

The second-order remainder shows what prediction error misses. The bias depends on how the errors in the two nuisance functions align, through the signed cross-product, and on the error in the exposure model; neither is visible in the marginal prediction errors. When a learner cannot represent a component shared by both nuisance functions, as OLS and GAM could not represent the interaction here, its errors align and produce bias even when each error is small. The remainder predicted the average bias well whenever the residuals were formed out of sample or the learner was too rigid to absorb noise in its own training data, but it explained little of the replicate-to-replicate variation in $\widehat\tau$, which is dominated by first-order sampling error.

The joint-error measure $\mathcal{D}_{\mathrm{joint}}$ was only weakly related to absolute estimation error. It is more useful as a description of how the two nuisance errors align than as a criterion for comparing causal estimators, and it should not be interpreted as a complete measure of the orthogonal-score remainder.

These results do not show that prediction error is irrelevant for causal inference. Prediction error remains a useful measure of nuisance-function accuracy, particularly for checking whether a learner estimates the underlying regression functions adequately. The limitation is that it does not, by itself, determine how nuisance estimation translates into the sampling distribution of the final causal estimator.

Several limitations should be noted. First, the simulation considered a constant treatment effect and a relatively simple partially linear causal structure, and the relationship between prediction error and causal performance may differ with treatment-effect heterogeneity. Second, the clustered analysis was limited to one cluster size and one set of variance components. Third, all methods used the same orthogonal score, so the results do not address estimators based on non-orthogonal moments; the difference between XGBoost and DML-XGBoost reflects cross-fitting, and the differences among OLS, GAM, and XGBoost reflect the learners. Fourth, XGBoost was used with fixed hyperparameters, and tuning could change the size of its own-observation term. Finally, the standard errors of both XGBoost estimators were 10\% to 16\% too small, so their coverage reflects variance estimation as well as bias. Variation in fitted nuisance functions across repeated training samples, which we did not evaluate, may carry information about causal performance beyond prediction error, particularly when examined across regions of the covariate distribution.

\section*{Data and Code Availability}

This study used simulated data generated from the data-generating mechanisms described in the Methods section. No human participant data were used. The simulation and analysis code are publicly available at \url{https://github.com/congca2/nuisance-function-prediction-causal-estimation}. Scenario~1 results were produced by \path{simulation_revised_v3.R}, Scenario~2 results by \path{simulation_revised_2_v3.R}, and Tables~\ref{tab:causal_performance} and~\ref{tab:prediction} and Figures~\ref{fig:prediction-error-causal-bias} and~\ref{fig:joint-error-causal-bias} by \path{make_tables_figures_v3.R} from the saved results.

\section*{Funding}

The author received no specific funding for this work.

\section*{Author Contributions}

Cong Cao: Conceptualization, Methodology, Formal analysis, Software, Visualization, Writing -- original draft, Writing -- review and editing.

\section*{Competing Interests}

The author declares no competing interests.

\bibliographystyle{plain}
\bibliography{ref}

\end{document}